\documentclass[11pt,a4paper]{article}
\usepackage[hyperref]{acl2021}
\usepackage{times}
\usepackage{latexsym}

\usepackage{microtype}

\usepackage{float}
\usepackage{epsfig}
\usepackage{graphicx}
\usepackage{amsmath}
\usepackage{amssymb}
\usepackage{import}
\usepackage{xifthen}
\usepackage{transparent}
\usepackage{svg}
\usepackage{array}
\newcolumntype{P}[1]{>{\centering\arraybackslash}p{#1}}
\usepackage{makecell}

\aclfinalcopy
\title{SMURF: SeMantic and linguistic UndeRstanding Fusion for Caption Evaluation via Typicality Analysis}

\author{Joshua Feinglass \textnormal{and} Yezhou Yang\\
  Arizona State University\\
  \texttt{\{joshua.feinglass,yz.yang\}@asu.edu} 
  }

\begin{document}
\maketitle
\begin{abstract}
The open-ended nature of visual captioning makes it a challenging area for evaluation. The majority of proposed models rely on specialized training to improve human-correlation, resulting in limited adoption, generalizability, and explainabilty. We introduce ``typicality'', a new formulation of evaluation rooted in information theory, which is uniquely suited for problems lacking a definite ground truth. Typicality serves as our framework to develop a novel semantic comparison, \text{SPARCS}, as well as referenceless fluency evaluation metrics. Over the course of our analysis, two separate dimensions of fluency naturally emerge: style, captured by metric \text{SPURTS}, and grammar, captured in the form of grammatical outlier penalties. Through extensive experiments and ablation studies on benchmark datasets, we show how these decomposed dimensions of semantics and fluency provide greater system-level insight into captioner differences. Our proposed metrics along with their combination, SMURF, achieve state-of-the-art correlation with human judgment when compared with other rule-based evaluation metrics\footnote{SMURF source codes and data will be released at \url{https://github.com/JoshuaFeinglass/SMURF}.}.
\end{abstract}

%%%%%%%%% BODY TEXT
\section{Introduction}
\begin{figure}[t]
\def\svgwidth{\columnwidth}
\includegraphics[width=\columnwidth,height=\textheight,keepaspectratio]{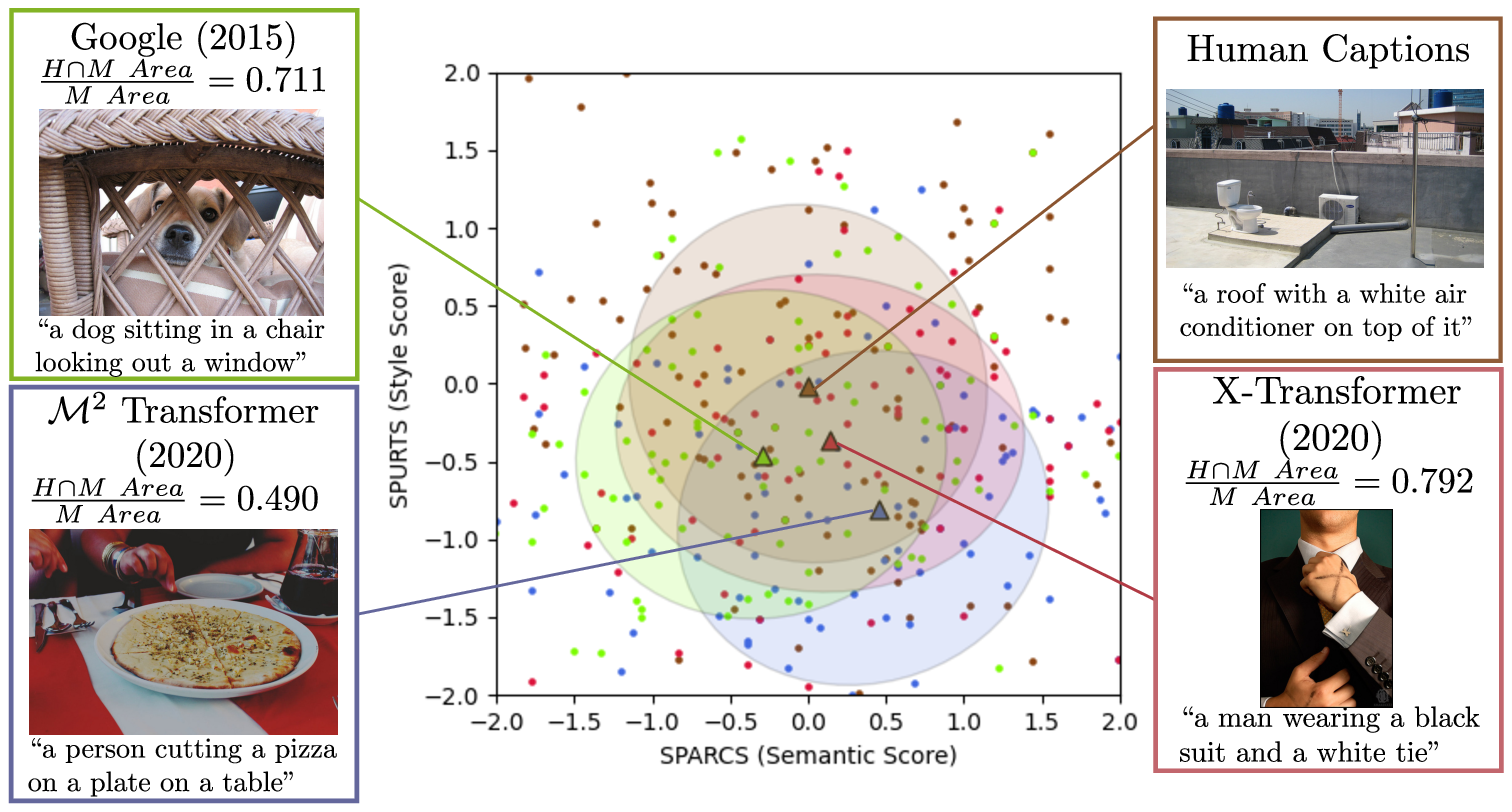}
\caption{Scatter plot utilizing standardizations of \text{SPARCS} and \text{SPURTS}. The ground truth captions are sourced from the Karpathy test split of the COCO dataset~\citep{coco2015,neuraltalk2015} with one used as a baseline for automatic captioners~\citep{meshed2020,x2020,google2015}. For each captioner, a 75\% confidence ellipse (1.15 standard deviations from the mean) is generated. A caption near the centroid of each captioner is shown as an example along with the caption scores from 100 randomly sampled images. The normalized ellipse overlap between an automatic captioner and human captions, $\frac{H \cap M\ Area}{M\ Area}$, gives an overall evaluation of typical performance at a system-level on a scale of 0 to 1, with 1 being human-caption level.}
\label{system-level}
\end{figure}
Visual captioning serves as a foundation for image/video understanding tools and relies on caption evaluation for identifying promising research directions. Rule-based caption evaluation approaches like the n-gram based CIDEr~\citep{cider2015} and parsed semantic proposal based SPICE~\citep{spice2016} specifically are able to provide researchers with meaningful feedback on what their algorithm is lacking. However, n-gram based methods are sensitive to stop words and sentence parsers are often inconsistent, leading to~\citet{spider2017} showing that neither method fully captures either the fluency or the semantic meaning of text. More recently proposed metrics attempt to learn cues of caption quality by training models via image grounding techniques~\citep{eval2018} or human and generated captions~\citep{bleurt2020}. These approaches, however, lack generality, require domain specific training, and offer little insight for improving captioners, leading to none of the proposed models being adopted for use as a caption evaluation benchmark. We instead postulate that quality in semantics and descriptive language is universally recognizable.
\par
The primary difficulty of caption evaluation is its cross-modal nature introducing ambiguity into the expected output, resulting in a ground truth that is no longer a single outcome, but a large set of potential outcomes of varying levels of quality. From this problem setting, the novel concept of ``typicality'' arises naturally. A desirable caption is one that is atypical enough linguistically that it uniquely describes the scene, follows typical natural language protocols, and matches a typical semantic description of a scene.
\par
Linguistically, the number of typical sequences is characterized by the entropy rate~\citep{infotheory}. Current work estimates the English language as having an entropy rate of only $1.44$ bits/letter~\citep{entropy2018}, implying that the typical set of English is only a tiny fraction of the full space of potential text. Self-attention transformers are language models that are able to identify the distinguishing contextual features of this typical set and as a result have now become the staple of natural language understanding tasks. Here we define typicality based on the distance of a candidate text's features from expected features of the typical set. We call this linguistic typicality estimation method Model-Integrated Meta-Analysis (\text{MIMA}) and use the function, $f_{\text{MIMA}}$, to create referenceless fluency metrics attune to captioning needs. Rather than assuming a predefined evaluation task and introducing bias by fine-tuning the self-attention transformer, our method extracts the inherent properties of language learned by transformers~\citep{bert2019,roberta2019} by treating self-attention layers as probability distributions as demonstrated in~\citet{analysis2019}. Our approach represents the first integration of a fluency specific metric that demonstrably improves correlation with human judgment for caption evaluation. 
\par
By removing stop words from the candidate text, $f_{\text{MIMA}}$ is able to create a metric that assesses a relatively new fluency criteria in captioning: style. We refer to this metric as Stochastic Process Understanding Rating using Typical Sets (\text{SPURTS}). Style can be thought of as the instantiation of diction and is necessary for generating human-level quality captions. Stylized captions describe a much smaller set of media, leading to machines instead generating the most typical caption that is still semantically correct. This results in a significant gap between machine and human captioners that can be seen in diction-based examples such as the use of the common words like ``dog'' and ``food'' instead of more descriptive words like ``Schnauzer'' and ``lasagna''. The other aspect of fluency assessed by $f_{\text{MIMA}}$ is grammar. Unlike style, grammar is not essential for caption quality, however, highly atypical syntax can potentially lead to awkward captions, so we develop a separate grammatical outlier penalty.
\par
We then define a lightweight and reliable typicality based semantic similarity measure, Semantic Proposal Alikeness Rating using Concept Similarity (\text{SPARCS}), which complements our referenceless metrics and grounds them to the reference captions. By matching word sequences, current methods limit the scope of their evaluation. Instead, we take non-stopword unigrams and further coalesce them into concepts through stemming, then combine the reference texts, like in~\citet{BSadjust2020}, using a novel semantic typicality measure of the reference text's concepts to evaluate the semantic similarity of a candidate and reference text.
\par
\text{SPURTS} and \text{SPARCS} can be used to assess system-level differences between captioners as shown in Figure~\ref{system-level}. Based on this analysis, the $M^2$ Transformer lags behind 2015 models in terms of similarity to human captions, even though both 2020 captioners achieved state-of-the-art results based on CIDEr standards. This difference becomes even more significant when you consider that the use of style makes it more difficult for a caption to be semantically correct. Human captions, $\mathcal{M}^2$ Transformer~\citep{meshed2020}, X-Transformer~\citep{x2020}, and Google~\citep{google2015} incur a total grammar outlier penalty of $-44.93$, $-7.47$, $-7.56$, and $-4.46$, respectively. In order to provide caption-level insight as well, we combine \text{SPURTS}, \text{SPARCS}, and our grammar outlier penalty into one metric - SeMantic and linguistic UndeRstanding Fusion (SMURF) - which rewards captions based on semantics and fluency. \\  
\textbf{Contributions:} Our key contributions are: \\
1. A novel and widely-applicable model meta-analysis technique, \text{MIMA}, which estimates the typicality of candidate text and which provides a means of assessing transformer robustness.\\
2. Three novel evaluation metrics useful for both caption-level and system-level evaluation: style-focused \text{SPURTS}, semantic-focused \text{SPARCS}, and their combination which incorporates grammatical outliers as well, SMURF. \\
3. Experiments showing that \text{SPARCS} and SMURF achieve SOTA performance in their respective areas of semantic evaluation and human-machine evaluation at both a system and caption-level.\\
4. Evidence showing that the performance of automatic evaluation metrics has been underestimated relative to voting-based human evaluation metrics.

\section{Related Work}
Originally, popular rule-based metrics from machine translation that were mostly n-gram based, namely METEOR~\citep{meteor2005}, BLEU~\citep{bleu2002}, and ROUGE~\citep{rouge2004}, were used for caption evaluation. ~\citet{cider2015} introduced the more semantically sensitive CIDEr which uses tf-idf to identify distinguishing n-grams and then compares them using cosine similarity. SPICE~\citep{spice2016} greatly improved upon n-gram based approaches by using a sentence parser to generate semantic propositions. Word moving distance scores~\citep{moverscore2019,wmd2016} have also been used for semantic evaluation with limited success. BERTScore~\citep{bertscore2019} used cosine similarity of embeddings from the self-attention transformer, BERT, and achieved state-of-the-art results on COCO but provided little interpretation of their approach.
\par
Domain specific training approaches have also been introduced with limited adoption.~\citet{eval2018,tiger2019,nneval2019} present a training approach for caption evaluation where an image grounding and/or caption based Turing test is learned based on training data from human and machine captioners. An adjusted BERTScore~\citep{BSadjust2020}, BLEURT~\citep{bleurt2020}, and NUBIA~\citep{trained2020} utilize transformer embeddings for comparison between reference and candidate text, then perform caption dataset specific fine-tuning of the model downstream.
\par
The importance of fluency in captioning has been widely recognized.~\citet{spider2017} attempted to integrate CIDEr and SPICE to create a cost function attune to both lexicographical and semantic qualities for captioning optimization. ~\citet{eval2018} identified the presence of less frequent, distinguishing words within human-generated text in the COCO dataset.~\citet{style2018} recognized the importance of style in captions and integrated it into their model without sacrificing semantics. 
\par
Referenceless evaluation, first proposed in ~\citet{grammar2016} as a referenceless grammar error correction (GEC) evaluation metric, has been recognized as an effective avenue for fluency evaluation as a whole~\citep{referenceless2017}, along with combined approaches~\citep{referenceless2018}. More recently, Perception Score~\citep{perception2020} outlined a general paradigm for training referenceless quality evaluation.\\
\section{Our Approach}
\subsection{Self-Attention Transformer Background}
\label{transformer}
First introduced in \citet{attention2017}, transformers are made of layers of parallel attention heads which extract contextual information about inputs using attention. They take in a sequence vector of tokenized words from candidate text, $y^n$, add start and separator/end tokens, and pass the input through a series of separate linear transforms with parameters, $p$, to create query, key, and value vectors, denoted as $q_i$,$k_i$,$v_i$, respectively. These vectors are then used to compute the attention weight parameters of the heads as shown:
\begin{align}
%\resizebox{.485 \textwidth}{!} 
\alpha_{ij}(y^n\!,p) &= \frac{exp(q_i^T k_j)}{\sum_{l=1}^n exp(q_i^T k_l)}, \\
o_i(y^n\!,p) &= \sum_{j=1}^n\alpha_{ij}v_{j},
\label{eq:1}
\end{align}
where $\alpha_{ij}$ and $o_{i}$ are each layer's attention weights and output, respectively. Here $\alpha_{ij}(y^n\!,p)$ is a joint distribution with marginal distributions $\alpha_i(y^n\!,p)=\sum_j \alpha_{ij}(y^n\!,p)$ and $\alpha_j(y^n\!,p)=\sum_i \alpha_{ij}(y^n\!,p)$.
%\par

BERT~\citep{bert2019} and RoBERTa~\citep{roberta2019} are encoder-decoder instantiations of transformers, pretrained on fundamental language tasks over large corpora. Both BERT and RoBERTa have achieved state-of-the-art results in various language understanding tasks. In order to speed up inference time, many papers have employed knowledge distillation to reduce the number of parameters these transformers require while still preserving their inference capabilities~\citep{distil2019,distilbert2019,distillinggen2020}.
\subsection{Information Theory Background}
\label{info_background}
Transformers like BERT and RoBERTa take text tokenized into sub-word components as input, capturing both the syntax and morphology of the text. The text sequences used as training data, $x^n$, can be modelled as a stationary ergodic stochastic process, ${\{X_k\}}_{k = 1}^{\infty}$, with instantiations limited to finite alphabet $\mathcal{X}$ and based on joint probability distribution, $P(X_1 = x_1, ..., X_n = x_n)$, whose transition predictability is governed by entropy rate, $H(\mathcal{X})$.
The entropy of a distribution, or entropy rate in the case of a stochastic process, can be used to describe the number of instantiations expected to be observed from a random variable or process, referred to as the typical set. From the Asymptotic Equipartition Property (AEP), it is known that the size of the typical set of sequences is bounded by
\begin{equation}
|A_n^{\epsilon}| \leq 2^{n(H(\mathcal{X})+\epsilon)},
\end{equation}
where $2^{nH(\mathcal{X})}$ estimates the size of the typical set.
\subsection{Model-Integrated Meta-Analysis}
\begin{figure}
\def\svgwidth{\columnwidth}
\includegraphics[width=\columnwidth,height=\textheight,keepaspectratio]{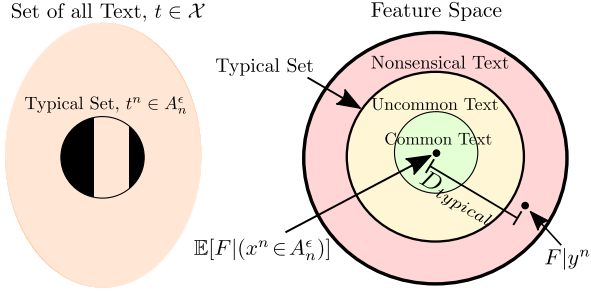}
\caption{Visualization of the typicality formulation introducing the concept of a typical set on the left and showing the distance proportional to typicality on the right.}
\label{typical_set}
\end{figure}
We assume that a self-attention transformer learns to fill in words from a sentence by extracting features, $F$. The quality of a piece of text can then be assessed by determining the distance of features taken by the model from candidate text, $Y^n=y^n$, from the expected value of features taken from correctly written text, $X^n=(x^n \! \in \! A_n^\epsilon)$, shown visually in Figure~\ref{typical_set} and mathematically in Equation~\ref{eq:dist}
\begin{equation}
D_{typical}=dist(F \ | \  y^n,\mathbb{E}[F \ | \  (x^n \! \in \! A_n^\epsilon)]).
\label{eq:dist}
\end{equation}
Here $dist$ does not does not refer to a specific distance metric and is instead an unspecified norm that exists in some realizable projection space. We then postulate the existence of a surrogate function, $f_{\text{MIMA}}$, which maps the sequence input and transformer parameter set, $p$, such that
\begin{equation} 
f_{\text{MIMA}}(y^n,p) \propto -D_{typical},
\end{equation}
resulting in a value indicating the typicality of a candidate input sequence. This value can be used to characterize the input for evaluation purposes.
\begin{figure}
\def\svgwidth{\columnwidth}
\includegraphics[width=\columnwidth,height=\textheight,keepaspectratio]{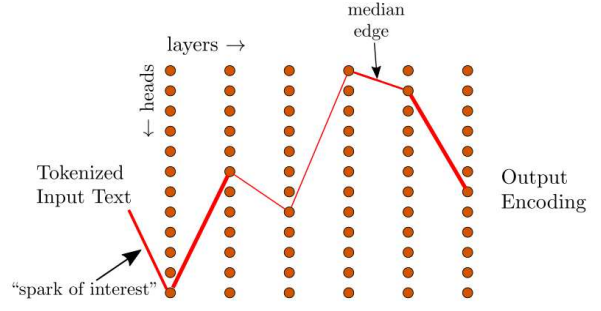}
\caption{Information flow used by $f_{\text{MIMA}}$ for estimating typicality of input in DistilBERT architecture.}
\label{info_flow}
\end{figure}
\par
\subsection{Attention-Based Information Flow as \text{MIMA} Function}
We postulate that input text that differs more greatly from members of the typical set generates a greater ``spark of interest'' in a transformer, resulting in greater information flow through parts of the network as shown in Figure~\ref{info_flow}. Conversely, if the input text is similar to the positive examples the transformer trains on, less information flows in through the layer, indicating that the model has already captured information about the sequence previously. We formulate information flow in terms of the attention dimensions $\alpha_i(y^n,p)$, $\alpha_j(y^n,p)$, and their joint distribution $\alpha_{ij}(y^n,p)$ as defined in Section~\ref{transformer}. We consider information flow based on the redundancy between $\alpha_i(y^n,p)$ and $\alpha_j(y^n,p)$ and use normalized mutual information (MI):
\begin{equation}
\resizebox{.48 \textwidth}{!}{
$\begin{aligned}
&I_{flow}(y^n,p)=MI\\
&=\frac{2\!*\!H(\alpha_i(y^n,p))+H(\alpha_j(y^n,p))-H(\alpha_{ij}(y^n,p))}{H(\alpha_i(y^n,p))+H(\alpha_j(y^n,p))},
\end{aligned}$
}
\end{equation}
as defined in~\citet{databook} to capture this redundancy.
\par
We are interested in attention heads with large information flow values, but find empirically that heads with the largest information flow values depend very little on the input and simply function as all-pass layers. Thus, we downselect to a single attention head information flow value to obtain
\begin{equation}
\begin{aligned}
&f_{\text{MIMA}}(y^n\!,p)\\
&=1-median_{layer}(max_{head}[I_{flow}(y^n\!,p)]).
\end{aligned}
\end{equation}
Here, the max over a given layer's attention heads captures the largest ``spark of interest''. The median removes outlier layers that have largely invariant information flow values.
\subsection{Caption Evaluation}
\text{MIMA} provides us with a foundation for computing the fluency of input text. We divide fluency into two categories: grammar and style. Grammar depends on the typicality of the sequence as a whole, $f_{\text{MIMA}}$, and is computed using the distilled BERT model since it achieves the highest Pearson correlation in the grammar experiment from Table~\ref{table:CoNLL}. Style depends on the distinctness, or atypicality, of the words directly associated with the image description, which we evaluate by removing the stop words from the text, then computing what we define as \text{SPURTS} as shown
\par
% It is not necessary for a caption to have any semantic similarity to the caption for a high syntactic score to be achieved. Stop words are a key example of this. The mutual information between the stop words and the caption is relatively small, but the stop words are essential for text with a small $D_{typical}$ value. Semantic evaluations fail to capture this
\begin{equation}
\text{SPURTS}=1-f_{\text{MIMA}}(y_{w\!/\!o},p),
\end{equation}
where $y_{w\!/\!o}$ is the candidate sequence without stop words and $f_{\text{MIMA}}$ is computed using the distilled RoBERTa model since it performs well on out-of-distribution text as shown in Figure~\ref{preliminary}.
\par
\begin{figure}
\def\svgwidth{\columnwidth}
\includegraphics[width=\columnwidth,height=\textheight,keepaspectratio]{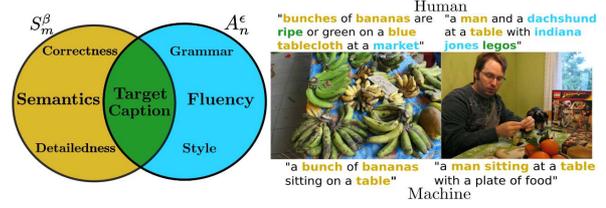}
\caption{Aspects of caption quality color-coded to corresponding words from evaluated COCO examples.}
\label{cap_quality}
\end{figure}

We formulate semantic similarity using typicality as well. Assuming a comprehensive set of all valid captions for a single image were available, we consider the distribution of all concepts, $\mathcal{S}$. Here we define concepts as the set stem terms that would remain if all stop words and affix/suffixes were removed from the text. The distribution of concepts sampled from such a set of captions, $S^m$, would have a typical set, $S_m^\beta$, of the most relevant concepts. Thus, a valid caption that is representative of the image semantically and demonstrates fluency should contain concepts that are members of the typical set of concepts, $S_m^\beta$, and be a member of the typical set of correctly formed language sequences defined in Section~\ref{info_background}, $A_n^\epsilon$, as shown in Figure~\ref{cap_quality}.
\par
To extract concepts from a caption, we use a stemmer on $y_{w/s}$ and estimate the typicality of each reference concept using the document frequency, $df$, of the concept across the available reference captions, $gt(\mathcal{S})$, where $gt$ is the function that maps concepts to a reference caption set. We then use an adjusted F1 score to determine the similarity between the reference concepts and candidate concepts. 
\par
The first portion of the F1 score is precision, corresponding to caption correctness. Our adjusted precision is
\begin{equation}
P(\mathcal{C},\mathcal{S})=\frac{\sum_{i}\frac{df_{gt(\mathcal{S})}(C_i)}{|gt(\mathcal{S})|}}{\sum_{i}(\frac{df_{gt(\mathcal{S})}(C_i)}{|gt(\mathcal{S})|}+\mathbb{I}[df_{gt(\mathcal{S})}(C_i)=0])},
\end{equation}
where $\mathcal{C}$ is the candidate concept set and $gt(\mathcal{S})$ is the reference caption set. Our approach equally weights correct and incorrect concepts if only one reference is used, but as the number increases, gradually decreases the importance of less common correct concepts.
\par
The second portion of the F1 score is recall, corresponding to caption detail. Our adjusted recall is
\begin{equation}
R(\mathcal{C},\mathcal{S})=\frac{\sum_{i}df_{gt(\mathcal{S})}(C_i)}{\sum_{i}df_{gt(\mathcal{S})}(S_i)}.
\end{equation}
where a candidate concept set, $\mathcal{C}$, which included all concepts from the reference set, $\mathcal{S}$, would achieve a score of 1.
\par
We then use the standard F1 score combination
\begin{equation}
\text{SPARCS}=F_1(\mathcal{C},\mathcal{S})=\frac{2*P(\mathcal{C},\mathcal{S})*R(\mathcal{C},\mathcal{S})}{P(\mathcal{C},\mathcal{S})+R(\mathcal{C},\mathcal{S})}.
\end{equation}
\par
To give an overall evaluation of performance, we fuse the proposed metrics. To begin, we standardize the output score distribution of human generated captions for each metric using the captions from the COCO Karpathy test split from Figure~\ref{system-level}, $metric'=\frac{metric-\mathbb{E}[metric(COCO_{test})]}{\sigma(metric(COCO_{test}))}$, creating $\text{SPARCS}'$, $\text{SPURTS}'$, and $f_{\text{MIMA}}'$. Utilizing the standardization, we use threshold, $T \! = \! -1.96$, corresponding to the left tail of a $95\%$ confidence interval, to represent the lower bound of expected human captioning performance. We then use $T$ to define a grammatical outlier penalty $G\!=min\!(\text{MIMA}'\!-\!T,0)$ and a style reward $D\!=\!max(\text{SPURTS}'\!-\!T,0)$. The quantities are combined as follows
\begin{equation}
%\begin{aligned}
\text{SMURF}
    = \begin{cases} 
      {\text{SPARCS}'+G}\ \ \  if\ \text{SPARCS}'<T, \\
      \text{SPARCS}'+D+G\ \ \ otherwise. \\
   \end{cases}
%\end{aligned}
\end{equation}
It can be interpreted as applying a semantic threshold, then incorporating the style reward since style is only beneficial for caption quality if the caption is semantically correct. For all of our proposed metrics, a larger value corresponds to higher quality caption.
\section{Experiments}
\subsection{Preliminary Experiment}
\begin{figure}
\centering
\def\svgwidth{\columnwidth}
\includegraphics[width=.98\columnwidth,height=.98\textheight,keepaspectratio]{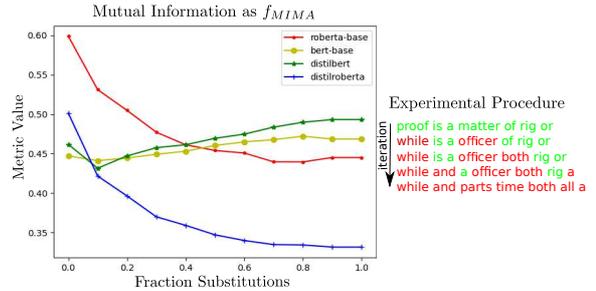}
\caption{Degradation iteration example and plot of each model's average $f_{\text{MIMA}}$ value as text degrades.}
\label{preliminary}
\end{figure}

We first seek to validate that our proposed $f_{\text{MIMA}}$, extracted from the attention layers of BERT, RoBERTa, and their knowledge distilled versions, is proportional to the distance from the expected value of features of the typical set. To this end, we create an experiment where we can control the randomness of input text. We begin with 11 different paragraphs from unrelated Wikipedia articles. We extract all the words from the paragraphs and create a word set corpus. We then sample 25 sentences from the paragraphs randomly. Each sentence is iteratively degraded by substituting a fraction of the words with random words from the word set corpus. At each iteration step, the sentences are passed through the transformers and the value of $f_{\text{MIMA}}$ is computed. Eventually the sentence is incoherent and bears no resemblance to ``natural'' text. The process and results can be seen in Figure~\ref{preliminary}. The average $f_{\text{MIMA}}$ value for our information flow formulation shows a strong correlation with the degradation in both models up until about 10\% of the tokens have been replaced, beyond which RoBERTa remains reliable but BERT does not, demonstrating RoBERTa's superior robustness.
\subsection{Datasets}
\noindent\textbf{CoNLL-2014} The CoNLL-2014 competition~\citep{conlltask2014} was a shared task of correcting grammatical errors of all types present in different sentences of an essay written by a learner of English as a second language. The essay consisted of 1312 separate sections to correct. A system-level human evaluation study of the grammatical quality of the corrected sentences from 12 competition submissions was presented in~\citet{humaneval2015}. Participants were asked to rate how natural the corrected sentences sounded and did not have access to any reference sentence.\\
\textbf{Microsoft COCO 2014} We use the Microsoft COCO validation set~\citep{coco2015}, comprised of 40,504 images, for a system-level human correlation experiment. These images are annotated with five human-generated captions, one of which is used as a baseline caption candidate. Human evaluations of competition entries were collected using Amazon Mechanical Turk (AMT). These evaluations were framed as questions from which 2 primary dimensions of system-level caption quality were derived as a ground truth to rank competitors: \textbf{M1} (percentage better than or equal to human description) and \textbf{M2} (percentage passing the Turing Test). Three additional categories were also included as an experimental ablation study but were not considered in the final competition ranking. In total, 255,000 evaluations were collected.\\
\textbf{Flickr 8K} We use the graded human quality scores for the 5,822 remapped captions from the Flickr 8k dataset~\citep{flickr2013} for a caption-level semantic human correlation study. The dataset was formed by selecting captions from one image and assigning them to another. These captions are then graded based on how well they align with the image using two different standards. The first standard is Expert Annotation, where human experts rate the image-caption pairing on a scale of 1 (caption and image unrelated) to 4 (caption describes image with no errors). Each caption-image pairing has 3 scores, which we combine by taking the average. The second standard is Crowd Flower Annotation, where at least 3 students vote yes or no on whether the caption and image are aligned.\\
\textbf{Composite Dataset} An additional dataset for caption-level study of semantic human correlation from~\citet{composite2015}. It contains 11,095 human judgments (on a scale of 1-5) over Flickr 8K, Flickr 30K~\citep{flickr30k2014}, and COCO and in contrast to the Flickr 8K dataset, includes machine generated captions in addition to human reference captions as candidates. Each evaluation is either based purely on correctness or detailedness.\\
\textbf{PASCAL-50S} Human evaluators were asked to identify which of two sentences, B or C, is more similar to reference sentence A. Unlike other caption datasets, human evaluators in Pascal-50S~\citep{cider2015} did not have access to the original image. The captions for sentence A were sourced from a 1000 image subset of the UIUC PASCAL Sentence Dataset~\citep{pascal2010} for which additional human captions were collected using AMT. Sentence B and C were sourced from both human and machine generated captions. The human captions were sourced from the original PASCAL dataset, resulting in four different pairing combinations: human-correct (HC), human-incorrect (HI), human-model (HM), and model-model (MM).
\subsection{System-Level Human Correlation}
System-level experiments evaluate how closely human evaluation and automatic evaluation models align in terms of their overall evaluation of captioning models. To confirm that $f_{\text{MIMA}}$ can capture grammar information, we replicate the experiment performed in ~\citet{grammar2016} and show improved performance over previous benchmarks in Table~\ref{table:CoNLL}. GLEU~\citep{gleu2015}, I-measure~\citep{imeasure2015}, and $\mathrm{M^2}$~\citep{msquared2012} are reference-based while their proposed ER, LT, and LFM are referenceless and based on linguistic features like $f_{\text{MIMA}}$. 
\begin{table}[h]
\centering % used for centering table
\footnotesize	
\begin{tabular}{l|l|l} % centered columns (4 columns)
%\begin{tabular}{p{2.5cm}|p{2.5cm}|p{2.5cm}}
%%\hline
%%\multicolumn{3}{|c|}{Country List} \\
\hline
%\thead{Attribute}&\thead{Primary \\ Application}&\thead{Context \\ Evaluation}&\thead{Semantic \\ Evaluation}&\thead{Method \\ Drawbacks}\\
%\hline
Metric & Spearman's $\rho$ & Pearson's $r$\\
\hline
GLEU & 0.852 & 0.838\\
ER & 0.852 & 0.829\\
LT & 0.808 & 0.811\\
I-measure & 0.769 & 0.753 \\
LFM & 0.780 & 0.742\\
$\mathrm{M^2}$ & 0.648 & 0.641\\
$\mathrm{BERT_{\text{MIMA}}}$ & 0.852 & \textbf{0.913}\\
$\mathrm{RoBERTa_{\text{MIMA}}}$ & \textbf{0.885} & 0.878\\
\hline
\end{tabular}
\vspace{1px}
\caption{CoNLL system-level human correlation experiment results utilizing distilled versions of BERT and RoBERTA.} % title of Table
\label{table:CoNLL} % is used to refer this table in the text
\end{table}
\par
We then benchmark our proposed caption evaluation metrics against the rule-based metrics used in the Microsoft COCO 2015 Captioning Competition, which still serve as the standard for caption evaluation, and the recall-idf configuration of BERTScore. We observe that the original COCO submissions and many of the original codebases for the submissions are not publicly available or do not provide pretrained models. Other authors attempt to reproduce the submissions using open source reimplementations that they have trained themselves, which will not be consistent with the submissions for which the human evaluations were performed. Thus, we instead opt to use the 4 representative baseline caption sets~\citep{google2015,montreal2016,neuraltalk2015} provided publicly by~\citet{eval2018}, which include 3 competition submissions from open sourced models and 1 human caption baseline. These are guaranteed to be consistent with their work and reproducible. In Table~\ref{table:COCO}, we show the COCO results for \text{SPARCS}, \text{SPURTS}, and SMURF.
\par
SMURF and BERTScore demonstrate the highest correlation with human judgment in this dataset. BERTScore's performance is partially due to incorporation of idf dataset priors also used by CIDEr, which we do not utilize to keep our metrics as general and consistent as possible. To illustrate this point, we also report BERTScore's correlation without idf weighting (BS-w/oidf) for this experiment. Despite its simplicity, \text{SPARCS} also performs well along with SPURTS. The rest of the metrics fail to adequately reflect human judgment. 
\begin{table}[h]
\centering % used for centering table
\footnotesize	
\begin{tabular}{P{0.175\columnwidth} | P{0.135\columnwidth} P{0.135\columnwidth} | P{0.135\columnwidth} P{0.135\columnwidth}} 
% \begin{tabular}{m{3cm} | m{1.5cm} m{1.5cm} | m{1.5cm} m{1.5cm}} 
% centered columns (4 columns)
%\begin{tabular}{p{2.5cm}|p{2.5cm}|p{2.5cm}}
%%\hline
%%\multicolumn{3}{|c|}{Country List} \\
\hline
%\thead{Attribute}&\thead{Primary \\ Application}&\thead{Context \\ Evaluation}&\thead{Semantic \\ Evaluation}&\thead{Method \\ Drawbacks}\\
%\hline
&\multicolumn{2}{c|}{M1}&\multicolumn{2}{c}{M2}\\ \cline{2-5}
& $\rho$ & $p$-value & $\rho$ & $p$-value\\
\hline
BERTScore & \textbf{0.986} & (0.014) & 0.985 & (0.015)\\
BS-w/oidf & 0.374 & (0.626) & 0.419 & (0.581)\\
Bleu-1 & -0.279 & (0.721) & -0.263 & (0.737)\\
Bleu-2 & -0.709 & (0.291) & -0.696 & (0.304)\\
Rouge-L & -0.812 & (0.188) & -0.802 & (0.198)\\
METEOR & 0.479 & (0.521) & 0.534 & (0.466)\\
CIDEr & 0.023 & (0.977) & 0.082 & (0.918)\\
SPICE & 0.956 & (0.044) & 0.973 & (0.027)\\
\text{SPARCS} & 0.874 & (0.126) & 0.894 & (0.106)\\
\text{SPURTS} & 0.956 & (0.044) & 0.955 & (0.045)\\
SMURF & 0.984 & (0.016) & \textbf{0.993} & (0.007)\\
\hline
\multicolumn{5}{l}{\textbf{M1:} Percentage of captions that are evaluated as better}\\
\multicolumn{5}{l}{or equal to human caption.}\\
\hline
\multicolumn{5}{l}{\textbf{M2:} Percentage of captions that pass the Turing Test.}\\
\hline
\end{tabular}
\vspace{1px}
\caption{Microsoft COCO system-level human correlation measured with Pearson's $r$ experiment results.} % title of Table
\label{table:COCO} % is used to refer this table in the text
\end{table}
\subsection{Caption-Level Human Correlation}
Caption level experiments evaluate how closely human evaluation and automatic evaluation models align for each individual caption. We begin with the Pascal-50S dataset in Table~\ref{table:pascal}. We follow the procedure used in~\citet{spice2016} and use the first 5 sentence A entries of each image. 
\par
The Pascal-50S dataset is based on a direct comparison between the reference and candidate captions, which gives similarity based metrics a distinct advantage. As a result, SPARCS achieves the top score in this experiment. Another interesting result is the fact that SPURTS performs reasonably well in the human-machine category despite having no access to the reference sentence. This shows SPURTS effectiveness as a Turing Test at both a system and caption-level, independent of semantic information. The additional information provided by \text{SPURTS} to SMURF in the human-machine category actually improves its performance.
\begin{table}[h]
\centering % used for centering table
\footnotesize	
\begin{tabular}{c|c|c|c|c|c}
\hline
Metric & HC & HI & HM & MM & All\\
\hline
BERTScore & 0.640 & 0.938 & \textbf{0.925} & 0.534 & 0.759\\
Bleu-1 & 0.619 & 0.903 & 0.883 & 0.555 & 0.740\\
Bleu-2 & 0.616 & 0.903 & 0.861 & 0.532 & 0.728\\
Rouge-L & 0.603 & 0.906 & 0.897 & 0.589 & 0.749\\
METEOR & 0.643 & 0.948 & 0.908 & 0.617 & 0.779\\
CIDEr & 0.633 & 0.949 & 0.866 & 0.639 & 0.772\\
SPICE & 0.628 & 0.938 & 0.866 & 0.637 & 0.767\\
\text{SPARCS} & \textbf{0.651} & \textbf{0.958} & 0.896 & \textbf{0.644} & \textbf{0.787}\\
\text{SPURTS} & 0.496 & 0.503 & 0.604 & 0.485 & 0.522\\
SMURF & 0.621 & 0.939 & 0.912 & 0.610 & 0.771\\
\hline
\end{tabular}
\vspace{1px}
\caption{PASCAL-50S caption-level classification accuracy for matching human evaluation results.} % title of Table
\label{table:pascal} % is used to refer this table in the text
\end{table}
\par
To evaluate our semantic metric specifically, we use the Flickr 8K and Composite dataset and follow the experiments specified in~\citet{spice2016}. However, we have discovered a flaw in previous comparisons between the correlation of automatic evaluation metrics with expert evaluation and inter-human correlation using the Flickr 8k dataset. Only a small subset of annotations between the Crowd Flower and Expert Annotations overlap, which often consists of ties causing the ranking metric to fail. To give a fair comparison, we also test the automatic metrics on a tie-free subset of the Flickr 8k data and use these results for human comparison. All of these results can be seen in Table~\ref{table:Flickr8k}. 
\par
\text{SPARCS} outperforms other metrics in the Flickr 8k dataset. However, SPICE outperforms \text{SPARCS} on the Composite dataset. This is likely due to the fact that evaluations of ``correctness'' in the Composite dataset are based on semantic propositions and do not consider partial correctness. 
\par
Additionally, these new results show that automatic metrics can actually outperform voting-based human metrics in terms of their correlation with experts, further motivating their use. This warrants further study as some recent datasets opt to use voting-based human metrics due to their ease of collection~\citep{crowd2019}.
\begin{table}[h]
\centering % used for centering table
\footnotesize	
\begin{tabular}{l|l|l|l} % centered columns (4 columns)
%\begin{tabular}{p{2.5cm}|p{2.5cm}|p{2.5cm}}
%%\hline
%%\multicolumn{3}{|c|}{Country List} \\
\hline
%\thead{Attribute}&\thead{Primary \\ Application}&\thead{Context \\ Evaluation}&\thead{Semantic \\ Evaluation}&\thead{Method \\ Drawbacks}\\
%\hline
Metric & Composite & Flickr 8K & Flickr Sub.\\
\hline
BERTScore & 0.388 & 0.362 & 0.530\\
Bleu-1 & 0.386 & 0.305 & 0.527\\
Bleu-2 & 0.394 & 0.316 & 0.577\\
Rouge-L & 0.393 & 0.277 & 0.511\\
METEOR & 0.404 & 0.411 & 0.611\\
CIDEr & 0.407 & 0.418 & 0.650\\
SPICE & \textbf{0.445} & 0.475 & 0.649\\
\text{SPARCS} & 0.431 & \textbf{0.481} & \textbf{0.716}\\
\hline
Inter-Human & - & - & 0.655\\
\hline
\end{tabular}
\vspace{1px}
\caption{Kendall's $\tau$ rank correlation with human judgment for the Flickr 8k and Composite datasets at a caption-level.} % title of Table
\label{table:Flickr8k} % is used to refer this table in the text
\end{table}
\par
\subsection{Generalization/Robustness Study}
\label{generalize_study}
We perform a caption-level generalizability and robustness case study on the most commonly used caption evaluation algorithms using the COCO validation set in Table~\ref{table:SPICE Failure}. We define a critical failure, $F$, as a disparity of greater than 1 between system-level human (M2) and caption-level algorithm correlation of a reference evaluation metric and a tested evaluation metric for a given caption set of an image. The last column of Table~5 shows the likelihood of a critical failure occurring for each metric.
\par
In a human study, we identify the primary cause of critical failure in the 20 most severe discrepancies in order to identify potential areas for improvement for each metric. We use SMURF as a reference evaluator for the other evaluators and SPICE as a reference for SMURF. The estimated probability of each of these failure causes is shown in the first three columns of Table~\ref{table:SPICE Failure}.
\par
The first failure cause, $c1$, refers to a scenario where the metric fails despite there being enough word overlap between the candidate and reference captions for a correct judgment to be made. This implies that the choice of words/sequences made by the metric for the comparison needs improvement. The second failure cause, $c2$, refers to the use of correct and distinct words or phrases by the human captioner that are not seen in the references. Lastly, we include the case where the reference evaluator may have incorrectly identified the correct caption ranking (according to the human annotator) as matching system-level human judgment. We refer to this as a reference failure, $RF$. 
\begin{table}[h]
\centering % used for centering table
\footnotesize	
\begin{tabular}{c|c|c|c|c} 
\hline

Metric & \scriptsize{$P(c_{1}|F)$} & \scriptsize{$P(c_{2}|F)$} & \scriptsize{$P(RF|F)$} & \scriptsize{$P(F)$}\\
\hline
CIDEr & 0.35 & 0.65 & 0.00 & 0.237 \\
METEOR & 0.65 & 0.35 & 0.00 & 0.205 \\
SPICE & 0.65 & 0.30 & 0.05 & 0.108 \\
SMURF & 0.40 & 0.30 & 0.30 & \textbf{0.038} \\
\hline
\end{tabular}
\vspace{1px}
\caption{Likelihood of critical failure and its causes.} % title of Table
\label{table:SPICE Failure} % is used to refer this table in the text
\end{table}
\par
The focus of previous studies has been robustness to distractors~\citep{nneval2019,eval2018,forced2016}. We observe no captions where this is a primary cause of failure. On the contrary, we find that each metric is highly susceptible to specific $c1$ scenarios: \\
\textbf{n-gram based}: Both CIDEr and METEOR are sensitive to stopwords, leading to rewards for words or sequences that supply no additional information.\\ 
\textbf{SPICE}:
Semantic proposal formation or sentence parsing issues can lead to the metric unpredictably failing to recognize highly informative proposals.\\
\textbf{SMURF}:
The metric may fail to adequately reward additional information if the words used are too common, like `few' or `some'.
\section{Conclusion and Future Work}
In this paper, we use information theory based typicality analysis to capture a new perspective on the problem of caption evaluation. Our analysis leads us to two caption evaluation metrics that capture separate dimensions of caption quality and a fused metric. We have performed experiments demonstrating their correlation with human judgment, showed how these methods could be used to perform multi-aspect system-level analysis of algorithm performance, and performed caption-level studies explaining why combining these two algorithms leads to more robust and generalizable evaluations. The underlying mechanism, \text{MIMA}, opens many new avenues for the analysis of self-attention transformers and potentially other models. Future work could also focus on optimal weighting between semantics and style.
\section{Ethical Impact}
Harmful bias, especially towards gender~\citep{women2018}, has been shown to be present in image caption datasets and is often further magnified by automatic captioners. Prior caption evaluation methods have the potential to further exacerbate the problem by rewarding such captions due to their reliance on dataset specific images or captions. Referenceless evaluations like our style metric, SPURTS, offer a preemptive approach for mitigating harmful dataset bias, like in Simpson’s Paradox~\citep{bias2019}, by utilizing intrinsic properties of descriptive language learned by self-attention models over far larger and more diverse corpora. This gives the evaluator a more wholistic view of caption quality rather than viewing the world through the lens of a single visual dataset
\section*{Acknowledgments}
{
The authors acknowledge support from the NSF Project VR-K \#1750082, the DARPA KAIROS program (LESTAT project), and the anonymous reviewers for their insightful discussion. Any opinions, findings, and conclusions in this publication are those of the authors and do not necessarily reflect the view of the funding agencies. 
}

\bibliographystyle{acl_natbib}
\begin{NoHyper}
\bibliography{anthology,acl2021}
\end{NoHyper}

\end{document}